%% file: main.tex
\documentclass[10pt,twocolumn,letterpaper]{article}

\usepackage{iccv}
\usepackage{times}
\usepackage{epsfig}
\usepackage{graphicx}
\usepackage{amsmath}
\usepackage{amssymb}

\usepackage{comment}
\usepackage{amsmath,amssymb} 
\usepackage{color}
\usepackage[symbol]{footmisc}
\usepackage{booktabs}
\usepackage{adjustbox}
\usepackage{hyperref}
\hypersetup{
  colorlinks   = true, 
  urlcolor     = red, 
  linkcolor    = blue, 
  citecolor   = green 
}

\usepackage{bm}
\usepackage{pifont}
\usepackage{xcolor}
\usepackage{array}
\usepackage{microtype}
\usepackage{wrapfig}

\newcommand{\cmark}{{\color{green}\ding{52}}}
\newcommand{\xmark}{{\color{red}\ding{56}}}
\newcolumntype{P}[1]{>{\centering\arraybackslash}p{#1}} 
\newcolumntype{M}[1]{>{\centering\arraybackslash}m{#1}} 

\def\eg{\emph{e.g}\onedot} 
\def\ie{\emph{i.e}\onedot}

\makeatother

\usepackage{animate}
\usepackage[font=small,labelfont=bf,tableposition=top]{caption}
\usepackage{dutchcal}

\usepackage{amssymb, amsmath, amsthm}
\usepackage{multirow}

\iccvfinalcopy 

\def\iccvPaperID{8612} 

\begin{document}

\title{UVA: Towards Unified Volumetric Avatar for View Synthesis, Pose rendering, Geometry and Texture Editing}

\author{
Jinlong Fan\\
The University of Sydney\\
{\tt\small jfan0939@uni.sydney.edu.au}
\and
Jing Zhang \\
The University of Sydney \\
{\tt\small chaimi.ustc@gmail.com}
\and
Dacheng Tao \\
The University of Sydney \\
{\tt\small dacheng.tao@gmail.com}
}


\maketitle


\input{fig_tex/first_fig}

\begin{abstract}

Neural radiance field (NeRF) has become a popular 3D representation method for human avatar reconstruction due to its high-quality rendering capabilities, e.g., regarding novel views and poses. However, previous methods for editing the geometry and appearance of the avatar only allow for global editing through body shape parameters and 2D texture maps. In this paper, we propose a new approach named \textbf{U}nified \textbf{V}olumetric \textbf{A}vatar (\textbf{UVA}) that enables local and independent editing of both geometry and texture, while retaining the ability to render novel views and poses. UVA transforms each observation point to a canonical space using a skinning motion field and represents geometry and texture in separate neural fields. Each field is composed of a set of structured latent codes that are attached to anchor nodes on a deformable mesh in canonical space and diffused into the entire space via interpolation, allowing for local editing. To address spatial ambiguity in code interpolation, we use a local signed height indicator. We also replace the view-dependent radiance color with a pose-dependent shading factor to better represent surface illumination in different poses. Experiments on multiple human avatars demonstrate that our UVA achieves competitive results in novel view synthesis and novel pose rendering while enabling local and independent editing of geometry and appearance. The source code will be released.

\end{abstract}

\section{Introduction}
Neural implicit fields are widely recognized for their remarkable success in 3D representation, owing to their high-quality rendering and flexible representation of complex shapes and scenes without the need for explicit surface meshes \cite{hoppe_MeshOptimization_1993}, voxels \cite{laine_EfficientSparseVoxel_2010, liu_NeuralSparseVoxel_2020, ashburner_VoxelbasedMorphometryMethods_2000}, or point clouds \cite{remondino_PointCloudSurface_2003}, especially after the emergence of NeRF \cite{mildenhallNeRFRepresentingScenes2020a}. Despite the numerous efforts made to adapt neural fields for dynamic objects \cite{pumarolaDnerfNeuralRadiance2021, parkNerfiesDeformableNeural2021, parkHyperNeRFHigherDimensionalRepresentation2021} in the wild scenes \cite{martin-brualla_NerfWildNeural_2021, sun_Neural3DReconstruction_2022, meshry_NeuralRerenderingWild_2019} and large scenes \cite{turki_MegaNeRFScalableConstruction_2022, tancik_BlocknerfScalableLarge_2022}, manipulating the reconstructed neural fields remains challenging due to the black-box nature of the implicit representation. Previous methods have attempted to disentangle lighting, materials, geometry, and/or texture as separate latent variables \cite{liuEditingConditionalRadiance2021}, neural texture maps \cite{xiangNeutexNeuralTexture2021}, or controllable sub-neural fields \cite{kastenLayeredNeuralAtlases2021} to facilitate the editing of the scene \cite{qiaoNeuPhysicsEditableNeural2022, lazovaControlnerfEditableFeature2023, zhengEditableNeRFEditingTopologically2022}. However, these approaches are limited to static objects or are only capable of global manipulation.

\input{tables/diff_check}
Editing the dynamic field becomes particularly challenging when dealing with articulated objects, \eg human avatars. Earlier works on reconstructing human avatars using neural radiance fields have mainly focused on novel view synthesis and novel pose rendering \cite{pengNeuralBodyImplicit2021, weng_HumannerfFreeviewpointRendering_2022}. With the given camera and human pose parameters, the reconstructed avatar can be rendered from different views and in various poses. Besides, by bounding the avatar with a 3D bounding box, it can be relocated to any position in the 3D world \cite{zhang_NeuVVNeuralVolumetric_2022}. However, these approaches still lack the ability to edit the geometry and texture of the avatar.

The other line of work stores appearance information as separate neural features, which can be decoded into textures via deferred rendering \cite{xiangNeutexNeuralTexture2021, shysheyaTexturedNeuralAvatars2019, zhiTexMeshReconstructingDetailed2020}. The correspondence between the 3D query points and the 2D neural texture map is determined using pre-defined UV mapping \cite{rajAnrArticulatedNeural2021, grigorevStylepeopleGenerativeModel2021, remelliDrivableVolumetricAvatars2022}. However, UV mapping is only defined in the 3D-to-2D direction, and mapping a specific pixel on the texture map to the 3D world is not directly available, which makes local texture editing difficult. Additionally, these methods can only control the shape of the human body globally via shape parameters in statistical human models, \eg SMPL \cite{loper_SMPLSkinnedMultiperson_2015}, which are unable to edit the geometry of the human body locally.

In general, editing the geometry of objects is more challenging than editing their texture, as changes to geometry can affect appearance, \eg, the length of limbs, whereas appearance can be updated while maintaining the same geometry. Since explicit mesh representations can be precisely controlled and edited by altering vertices and faces, some approaches transfer a pre-trained neural radiance field to a polygon mesh by baking the geometry and appearance information from coordinate-based networks \cite{yangNeuMeshLearningDisentangled2022, xu_DeformingRadianceFields_2022}. 
However, utilizing two-stage distilling methods poses a challenging task, as it requires manual and extensive parameter tuning while being only suitable for static objects. How to edit the geometry of dynamic avatars remains an open question.

In this paper, we present a novel approach that enables both geometry deformation and local texture manipulation, while retaining the ability to render novel views and poses. The core idea is to encode geometry and appearance in separate fields and anchor the neural field on a deformable mesh in canonical space. Technically, we first transform the query points, which are sampled from variant poses in each frame, to the canonical space via the skinning motion field. The geometry and appearance fields are then represented by two sets of structured latent codes independently, which are attached to anchor nodes on the canonical mesh. Since the latent codes are tightly aligned with the mesh nodes, the deformation of the mesh can directly guide the deformation of the neural field. Meanwhile, the texture can also be edited locally, as each of the latent codes can be tuned individually. 
Although the discrete latent codes defined on mesh nodes can be diffused to the entire space using nearest-neighbor (NN) interpolation, it results in spatial query indistinguishability. To address this, we propose using a local signed height indicator comprising UV coordinates and point-to-surface distance to mark query points. The local signed height, relative to the anchored mesh, can track mesh deformation.

Thanks to the mesh-guided geometry and appearance, as well as mesh-tracked signed height, our proposed model, named \textbf{U}nified \textbf{V}olumetric \textbf{A}vatar (\textbf{UVA}), offers an all-in-one solution for novel view synthesis, novel pose rendering, local geometry deformation, and local texture manipulation. The comparison against existing methods regarding the rendering and editing abilities is shown in Table \ref{tab:check_check}. In summary, the main contribution of this paper is threefold:
\begin{itemize}
    \item We propose a novel baseline model named Unified Volumetric Avatar that enables local and independent editing of both geometry and texture while retaining the ability to render novel views and poses.
    \item We propose to use a skinning field to represent the motion while employing structured latent codes and the signed height indicator to characterize the neural field. Both are defined relative to the anchored mesh, facilitating the tracking of mesh deformation.
    \item The experimental results demonstrate the efficacy and versatility of our UVA, establishing it as a baseline model capable of rendering and manipulation.

\end{itemize}

\input{fig_tex/overview}

\vspace{-2mm}
\section{Related Work}
\paragraph{Neural human.}
NeRF has emerged as an effective approach for volume rendering, enabling high-quality novel view synthesis and serving as a compact 3D representation in various applications \cite{mildenhallNeRFRepresentingScenes2020a}. While neural rendering originally focused on static scenes or objects, recent works have introduced a deformation field to adapt the approach to dynamic scenes or objects, particularly for human reconstruction. Typically, a motion field such as a displacement field \cite{parkNerfiesDeformableNeural2021, parkHyperNeRFHigherDimensionalRepresentation2021} or a skinning field based on human prior \cite{pengAnimatableNeuralRadiance2021} is used to align the dynamic human with a template in canonical space, where the canonical radiance field captures the human in coordinate-based networks. However, implicit representations of reconstructed humans do not allow for easy manipulation. Our method employs structured latent codes and a signed height indicator to describe the canonical space based on mesh-based representation, enabling simple and intuitive manipulation of the reconstructed human.

\paragraph{Editable neural field.}
To make the neural field editable, several previous works have proposed various approaches. Liu et al. \cite{liuEditingConditionalRadiance2021} introduce learnable variables to control the attributes of the object, while Xiang et al. \cite{xiangNeutexNeuralTexture2021} disentangle textures from the geometry as separated neural texture maps for texture editing. Yuan et al. \cite{yuanNeRFeditingGeometryEditing2022} build a continuous deformation field between the edited mesh and the reconstructed mesh and use it to render the edited object. On the other hand, Xu et al. \cite{xu_DeformingRadianceFields_2022} and Yang et al. \cite{yangNeuMeshLearningDisentangled2022} distill the neural field to a controllable polygon mesh for subsequent editing. However, the two-stage distilling methods are difficult to use as it requires extensive parameter tuning. In contrast, our method is end-to-end trainable, and the canonical neural field is jointly optimized with the skinning field from scratch, requiring no manual effort.

\paragraph{Editable Human Avatar.}

Editing a human avatar is a more challenging task than editing the neural field of a static object because the avatar must maintain its drivability and natural appearance after modifications. Conventional techniques typically encode the avatar's texture as a feature map and points on the surface of the 3D object are projected onto the map using precomputed 3D-to-2D UV mapping to extract its color feature.  \cite{zhangNDFNeuralDeformable2022, remelliDrivableVolumetricAvatars2022, alldieckTex2shapeDetailedFull2019, shysheyaTexturedNeuralAvatars2019, zhiTexMeshReconstructingDetailed2020}. However, direct editing of neural features can be challenging. To address this issue, \cite{maNeuralParameterizationDynamic2022} and \cite{chenUVVolumesRealtime2022} combine explicit RGB texture stacks with implicit neural texture maps, allowing direct editing of RGB textures. Nevertheless, these methods lack the capacity for geometry editing. In contrast, our proposed method allows for the local and independent manipulation of both the texture and geometry of the reconstructed human avatar.

\section{Method}
Figure \ref{fig:overview} provides an overview of our proposed method, which comprises two key components: a backward skinning motion field (Sec. \ref{sec:motion}) and a canonical radiance field (Sec. \ref{sec:canonical}). Given a query point $\mathbf{x}_t$ in the observation space, we first apply the skinning motion to deform it into the canonical space. Here, we combine the skeleton motion with a non-rigid deformation field as the skinning field. In the canonical space, we represent the geometry field and texture field using two sets of structured latent codes that are anchored on a deformable mesh. To capture pose-dependent surface illumination variance, we modulate the radiance color with a pose-conditioned shading factor. A key advantage of our approach is that the geometry and texture are stored in two separate and structured fields, which enables independent and local editing of both geometry (Sec. \ref{sec:geo_edit}) and texture (Sec. \ref{sec:app_edit}).

\subsection{Skinning Motion}
\label{sec:motion}
Parametric human body models, such as SMPL  \cite{loper_SMPLSkinnedMultiperson_2015} or SMPLx \cite{pavlakos_ExpressiveBodyCapture_2019} model , typically use linear blending skinning (LBS) to deform points from the canonical pose to the target pose based on rigid bone transformations and skinning weights \cite{kavan_SkinningDualQuaternions_2007, james_SkinningMeshAnimations_2005}. However, these skinning weights are only defined on the template mesh points. Recent methods have used neural networks to predict the spatial skinning weights for arbitrary 3D points in space \cite{liTavaTemplatefreeAnimatable2022, wangArahAnimatableVolume2022}. While implicit LBS weights can provide detailed deformation, joint training of the radiance field and skinning field can result in a local optimum \cite{pengAnimatableNeuralRadiance2021}. In this paper, we employ the nearest neighbor (NN) interpolation method to diffuse the skinning weights into the space. The NN motion field is a simple yet effective technique for human deformation, and it helps to stabilize the training of the separated radiance fields. Moreover, the NN skinning field is naturally a mesh-based representation, making it possible to track vertex movements (\ie, deformation of the mesh) and thus desired for geometry editing.

Here, we take SMPL as the parametric model. For a specific point $\mathbf{x}_t$ in target pose $\mathbf{\theta}_t$, we first determine its Top-K nearest SMPL mesh vertexes $\{\mathbf{v}_i\}_{i=1}^K$. Given the LBS weights $\mathbf{w}^b_i$ of each vertex ${\mathbf{v}_i}$, we interpolate the LBS weights using the inverse distance $w_i=1/ \left \| \mathbf{x}_t - \mathbf{v}_i \right \| _2$:
\begin{equation}
    \mathbf{w}^b = \sum_{i=1}^K {\mathbf{w}^b_i w_i}, w_i=\frac{w_i}{\sum w_i}.
    \label{lbsw}
\end{equation}
Using this NN skinning motion field, we can deform $\mathbf{x}_t$ as:
\begin{equation}
    LBS^{-1}(\mathbf{x}_t) = \left(\sum_{b=1}^{N+1} \mathbf{w}^b \mathbf{B}^b \right)^{-1} \mathbf{x}_t, 
\end{equation}
where $\mathbf{B}^b$ denotes the rigid transformation of bone $b$, and $N$ represents the total number of bones. In order to more effectively describe the motion of off-the-body points, an additional transformation is used for the background points \cite{weng_HumannerfFreeviewpointRendering_2022}. Specifically, if the distance between a given point and its nearest SMPL vertex exceeds a certain threshold, we classify the point as a static background point, assign it a bone weight of zero, and set the background weight to one.

LBS only accounts for rigid transformations. To accommodate non-rigid deformation, we make use of a neural network, denoted by $\mathcal{F}_{\Delta}$, to predict per-point displacement $\Delta$ in canonical space. As point movements are inherently tied to pose, we condition the deformation on the target pose $\mathbf{\theta}_t$ via the following equation:
\begin{equation}
    \Delta = \mathcal{F}_{\Delta}(\phi(\mathbf{\Vec{h}}), \mathbf{\theta}_t),
    \label{eq: delta}
\end{equation}
where $\phi(\cdot)$ represents the position encoding function \cite{tancikFourierFeaturesLet2020}, and $\mathbf{\Vec{h}}$ is the signed height indicator, which will be introduced in Sec. \ref{sec:canonical}. Combining this neural deformation field with the NN skinning field, we can obtain canonical points $\mathbf{x}_c$ as:
\begin{equation}
    \mathbf{x}_c = LBS^{-1}(\mathbf{x}_t) + \Delta.
\end{equation}

\subsection{Canonical Space}
\label{sec:canonical}
We represent the volumetric avatar's geometry and appearance using two separate neural radiance fields in canonical space. Each field is composed of a set of structured latent codes $l$ which are anchored to a deformable mesh \cite{pengNeuralBodyImplicit2021, zhengStructuredLocalRadiance2022, kwonNeuralHumanPerformer2021, chen_GeometryguidedProgressiveNerf_2022}. When the mesh is deformed, the position of the structured latent codes changes correspondingly, resulting in deformed neural fields. Local appearance latent codes in the region of interest can be adjusted with new patterns, allowing targeted manipulation of appearance while leaving other areas unaffected. In this way, it enables independent and local editing of either the geometry or appearance of the reconstructed volumetric avatar.

\paragraph{Anchor mesh.}
In order to edit the geometry and appearance locally, we employ a deformable mesh $\mathcal{M}=\{\mathbf{v}_i\}$ as an anchor. In principle, the anchored mesh can take any form, such as a parametric model or one obtained from off-the-shelf methods \cite{dong_PINALearningPersonalized_2022, xiu_ICONImplicitClothed_2022, jiang_SelfReconSelfReconstruction_2022}. Here, we use the SMPL mesh in \textit{rest} pose as the anchored mesh for our baseline. The structured latent codes $l_{geo}$  and $l_{rgb}$ are stored at the anchor nodes on the mesh, where the mesh vertices serve as the anchor nodes for simplicity, though other sampled nodes may also be used \cite{yangNeuMeshLearningDisentangled2022, zhengStructuredLocalRadiance2022}. For a 3D point $\mathbf{x}$ in canonical space, we use NN interpolation to get its latent code, similar to the way we obtain the LBS weights:
\begin{equation}
    l_{geo/rgb} = \sum_{i=1}^K l_{geo/rgb}^i w_i,
\end{equation}
where $w_i$ is the inverse distance weighting defined in Eq. \ref{lbsw}. However, such interpolation alone can lead to ambiguity for points along the direction perpendicular to the surface, where multiple points can have the same latent codes but in different positions. To address this, we introduce a local signed height $\mathbf{\Vec{h}}$ as an indicator \cite{pengSelfNeRFFastTraining2022, liuNeuralActorNeural2021, zhangNDFNeuralDeformable2022}. The signed height plays a similar role as the signed distance but includes the UV coordinate of $\mathbf{x}$. As the body surface is intrinsically a 2D manifold, and the anchored mesh can be projected onto a pre-defined 2D UV map, the signed height indicator can locate any arbitrary point in 3D to a 2.5D volume by equipping the UV coordinate with a signed distance.
\begin{equation}
    \mathbf{\Vec{h}} = [u, v, \mathbf{\Bar{d}}],   
\end{equation}
where $[u, v]$ is the 2D coordinate of $\mathbf{x}$ on the UV map, $\mathbf{\Bar{d}}$ is the signed distance. To calculate $\mathbf{\Bar{d}}$, we first project $\mathbf{x}$ onto the mesh surface at $\mathbf{x}_0$ and obtain the direction of point $\mathbf{x}$ $\mathbf{\Vec{r}} = \mathbf{x} - \mathbf{x}_0$. Given the surface normal $\mathbf{\Vec{n}}$ at $\mathbf{x}_0$, the signed distance can be calculated as $\mathbf{\Bar{d}} = sign(\mathbf{\Vec{r}} \cdot \mathbf{\Vec{n}}) \| \mathbf{\Vec{r}} \|_2$.

\paragraph{Mesh-based geometry.}
In previous methods such as \cite{xu_DeformingRadianceFields_2022} and \cite{yangNeuMeshLearningDisentangled2022}, the polygon mesh is baked from a pre-trained neural field and then taken as a proxy for user editing. However, the two-stage distilling methods are difficult to use as it requires manual and extensive parameter tuning. In contrast, our method jointly optimizes the geometry field with other components from scratch without any extra effort. Given the interpolated geometry latent code $l_{geo}$ and the signed height $\mathbf{\Vec{h}}$, we use an MLP $\mathcal{F}_{\sigma}$ to predict the density $\sigma$ as follows:
\begin{equation}
    \sigma = \mathcal{F}_{\sigma} (\phi(\mathbf{\Vec{h}}), l_{geo}),
\end{equation}
where $\phi(\cdot)$ is the same position encoding function as in Eq. \ref{eq: delta}. As the structured latent codes and the signed height indicator are both tightly attached to the canonical mesh, the density field has the ability to track the deformation of the anchored mesh, which is crucial for mesh-guided geometry editing.

\paragraph{Mesh-based appearance.}
In NeRF, the radiance field is conditioned on ray directions. However, when dealing with posed humans, the target rays are bent in canonical space, and the sampled points along a ray no longer share the same view directions. Additionally, self-occlusion can result in points along the same ray having different illumination. To address these issues, we replace the ray direction-dependent radiance color with a pose-related shading factor. Specifically, we decompose the radiance color in canonical space into two factors. The first factor comprises pose-independent RGB colors that are shared across all frames. We utilize a neural network $\mathcal{F}_{c}$ that takes the signed height and appearance latent code as input to predict the shared RGB colors:
\begin{equation}
    \mathbf{c}_0, \mathbf{f} = \mathcal{F}_{c} (\phi(\mathbf{\Vec{h}}), l_{rgb}), 
\end{equation}
where $\mathbf{c}_0$ represents the RGB color in canonical pose and $\mathbf{f}$ is the output feature vector. The second factor is a pose-dependent shading factor that differs from frame to frame. Conditioned on the target pose $\mathbf{\theta}_t$, we use a shallow MLP $\mathcal{F}_s$ to estimate the per-point shading factor $s$ for each frame:
\begin{equation}
    s = \mathcal{F}_{s} (\mathbf{f}, \mathbf{\theta}_t).
\end{equation}
For points in the target pose, we modulate $\mathbf{c}_0$ with the shading factor to obtain the final color $\mathbf{c} = \mathbf{c}_0 \odot s$.

\input{fig_tex/geo_edit_313}

\subsection{Mesh-guided Geometry Editing}
\label{sec:geo_edit}
Traditional parametric body models, such as SMPL, offer global control over the shape of the body through shape parameters. However, these parameters affect the entire mesh, and any changes to them will alter the whole body. Our proposed method, UVA, offers local geometry editing in fine detail, allowing for precise modifications such as bulges on specific parts of the body, by attaching structured latent codes to anchor nodes on the mesh and defining signed height indicators relative to the mesh surface. This ensures that the density field can track the movement of anchored nodes and the deformation of the mesh, enabling mesh-guided geometry editing. Users can move anchor nodes easily and freely in 3D modeling software (\eg, Blender) or with out-of-the-box mesh deforming methods (\eg, as-rigid-as-possible (ARAP) \cite{sorkine_AsrigidaspossibleSurfaceModeling_2007}). Additionally, the edited avatar can be rendered in novel views and poses, as our skinning motion field is also mesh-based. By copying the movements of corresponding vertices from the canonical mesh to the target mesh, the skinning field can be deformed in sync, enabling animation.

\subsection{Texture Editing}
\label{sec:app_edit}
Texture editing for human avatars is still a challenging task in the field. In previous methods, attempts have been made to disentangle the geometry and appearance into separate neural implicit fields, and update the entire texture with new ones by means of neural texture map replacement \cite{xiangNeutexNeuralTexture2021}, or by painting on the 2D neural texture stack \cite{chenUVVolumesRealtime2022, maNeuralParameterizationDynamic2022}. In contrast, our proposed method represents the appearance field with anchored structured latent code, which allows for free 3D editing without requiring expert knowledge to catch the 2D-to-3D correspondence.

Geometry editing focuses on manipulating the positions of anchor nodes, while texture editing involves modifying the underlying latent code itself. It is possible to replace or fine-tune the latent codes with the desired texture, and local edits to the latent codes associated with interested anchor nodes will not affect the other regions or the geometry field, \ie, allowing users to edit the texture locally and independently.
In principle, there are no limitations to the types of texture editing that can be performed. Here we take texture swapping and texture painting as two examples.

\paragraph{Texture swapping.}
Given two trained UVAs, we can swap their textures by selecting anchor nodes of interest in 3D or by projecting 2D mask rays into 3D space to identify the closest nodes to be considered. Similar to \cite{yangNeuMeshLearningDisentangled2022}, non-rigid 3D alignment is performed to align the source and target anchor nodes. Intuitively, there are two ways to infer the swapping texture. One way is to assign the NN interpolated source codes to the corresponding target ones and decode the pre-blended codes with the source decoder. The other way is to infer the source texture in the target pose and blend the pre-inferred texture in the swapping area. We chose the pre-blended method due to its better 3D consistency.

\paragraph{Texture painting.}
Given an arbitrary painting and a binary mask, we identify the affected anchor nodes using the point-to-ray distance. To obtain a smooth texture on the boundary, we dilate the mask slightly and use it to sample the training rays, following \cite{yangNeuMeshLearningDisentangled2022}. These locally selected latent codes are fine-tuned to match the painting, which transfers the 2D painting to 3D structured latent codes and the edited texture can be rendered in novel views and poses. However, fine-tuning the texture decoder locally is non-trivial. In UVA, a pixel value is uniquely determined by the interpolated latent code, the local signed height indicator, and the shading factor, among which only the signed height is not trainable. We can fix the shading network and the uninvolved latent codes, but the texture decoder may overfit to the signed height during single-image fine-tuning and generate artifacts at other positions. During training, to prevent the signed height from dominating the training and to make the network depend only on spatial position, we use a higher learning rate for the latent code (\eg, 10 times higher in our experiments). In texture painting, we further increase the learning rate gap to reduce the impact of the signed height indicator. With spatial-aware fine-tuning, the texture can be edited locally and independently.

\section{Experiments}
\subsection{Experiment Settings}
\textbf{Dataset and metrics.} 
To demonstrate the effectiveness of our proposed method, we conduct experiments on nine performers from the ZJU-MoCap dataset \cite{pengNeuralBodyImplicit2021}. We choose ZJU-MoCap as our testbed due to the sprawling pose and relatively slow motion of the performers in this dataset. Additionally, considering the simplicity of our NN skinning field, we believe that evaluating UVA on ZJU-MoCap is reasonable. The ZJU-MoCap setup includes 23 cameras, and following \cite{wangArahAnimatableVolume2022}, we select four equally spaced views and choose frames ranging from 60 to 300 for training, while the remaining frames are used for evaluation. In novel pose rendering, unseen poses in the test set are set as the target poses and images are rendered from test views. For geometry and appearance editing, we render the avatar in both novel views and novel poses. To compare our method with existing methods, we use standard metrics such as LPIPS \cite{zhang_UnreasonableEffectivenessDeep_2018}, SSIM, and PSNR. The results of more performers can be found in supplementary materials.

\textbf{Implementation details.} 
The network is optimized using Adam \cite{kingma_AdamMethodStochastic_2015} with a learning rate decay from $5e^{-4}$ to $5e^{-6}$, and a $10\times$ learning rate for the structured latent codes. The dimensions of geometry latent codes and appearance codes are both 32. The appearance network consists of an 8-layer MLP with a skip connection at the 4th layer. The density network is a 4-layer MLP with softplus activation, while the shading network is a 3-layer MLP. The motion field and canonical field are jointly trained for 300K iterations with a batch size of 1024. The training process takes about 10 hours using two NVIDIA Tesla V100 GPUs.

\subsection{Novel view and Novel pose synthesis}
For novel view synthesis and novel pose rendering, we compare our method against five existing approaches: 1) Neural Body (NB) \cite{pengNeuralBodyImplicit2021} diffuses per-SMPL-vertex latent codes in observation space to condition the NeRF model and achieves high-quality novel view synthesis results on training poses; 2) Ani-NeRF \cite{pengAnimatableNeuralRadiance2021} learns a backward LBS weight field and a canonical NeRF to reconstruct the human avatar; 3) A-NeRF \cite{suAnerfArticulatedNeural2021} employs skeleton-relative embedding to predict the radiance field; 4) we implement TAVA-NN based on TAVA \cite{liTavaTemplatefreeAnimatable2022}, but replace the learnable deformation field with the NN skinning field to verify the effectiveness of our canonical field representation; 5) ARAH \cite{wangArahAnimatableVolume2022} adopts a joint root-finding module to establish the correspondence between the observation space and the canonical space and stores the geometry in SDF. It performs very well on out-of-distribution poses for novel view and novel pose rendering.

\textbf{Quantitative results} are presented in Table~\ref{tab:aver_metrics}. Our method exhibits competitive performance on novel view synthesis and comparable results on novel pose synthesis. Notably, UVA outperforms A-NeRF on novel pose rendering. While A-NeRF utilizes over-parameterized bone-relative embeddings to locate 3D query points, our method leverages mesh-relative signed height indicators, demonstrating its superior ability. Compared with Ani-NeRF, our method performs marginally better, showing that the NN skinning field has a comparable ability to the neural deformation field. Although TAVA-NN employs the same motion field as our method, our approach yields slightly better results, verifying the effectiveness of our canonical field.

\textbf{Qualitative results} are shown in Figure \ref{fig:nv_zju} and Figure \ref{fig:np_zju}. As can be seen, despite using a simple and intuitive approach (NN interpolation) to diffuse the LBS weights and latent codes into the continuous space, UVA achieves comparable performance as Ani-NeRF that uses the learnable deformation field and NB that employs the 3D sparse convolution diffusion. Moreover, our method shows good generalization to unseen poses, while A-NeRF without surface priors struggles in this regard. We attribute this to the mesh-guided representations and surface-relative signed height indicator.

\input{tables/nv_np_zju_avr}

\input{fig_tex/nv_zju}
\input{fig_tex/np_zju}

\input{fig_tex/edit}

\subsection{Geometry Editing}
The proposed mesh-guided geometry neural field allows us to track the deformation of the anchored mesh and animate it with novel poses. In Figure \ref{fig:geo_editing}, we show the rendered results of edited geometry in several poses. The first row shows the rendered results of the reconstructed avatar. In the second row, the geometry is globally edited by adjusting all vertex positions on the mesh via the SMPL shape parameters, where the structured latent codes move accordingly, and the geometry field deforms as expected. In the third and last row, we show UVA's ability to deform the field locally. We import the anchored mesh into a 3D graphics tool, \eg, Blender, and edit the mesh freely. In the third row, we edit the geometry by changing the size of the limbs, and in the last row, we stretch parts of the face, ears, and arms (Although these manipulations may appear simplistic, they serve to demonstrate UVA's capability for local editing). UVA leverages a mesh-guided geometry field and mesh-based signed height indicator to yield reasonable global manipulation and local deformation outcomes. The user-friendly editing process provides a ``What You See Is What You Get'' experience. 

\subsection{Texture Editing}
We demonstrate the ability of our method to perform texture editing through texture swapping and texture painting in Figure \ref{fig:editing}. In texture swapping, we swap the latent codes and texture decoder in the target area to generate the corresponding color. To smooth the texture near the boundary after editing, we blend the latent codes across target areas with their neighboring codes. Here, we swap the lower half of the body of two subjects, 313 and 394, with manually masked-out target areas. In texture painting, we update the local latent codes of interest through per-vertex fine-tuning to match the target texture, while the rest of the latent codes remain fixed. In Figure \ref{fig:editing}(b), we fill the area of interest with a reference image (\ie, the Lena image), and in Figure \ref{fig:editing}(c), we draw a pattern ``H'' on the T-shirt. The visual results demonstrate that through the spatial-aware and per-latent code fine-tuning, the texture is edited locally without affecting the non-interested areas and is independent of the manipulation of geometry.

\subsection{Ablation Study}
Table \ref{tab:ablation} shows the ablation study results on the key components of UVA using the sequence of subject 313 for novel view synthesis and novel pose rendering.

\textbf{Displacement field.} 
For novel view synthesis, the target pose is included in the training poses, and therefore, the displacement $\Delta$ has a marginal effect on point alignment. However, for novel pose rendering, unseen poses lead to frame-by-frame deformation that cannot be captured by skeleton motion alone. In such scenarios, the displacement field $\Delta$ matters for achieving better rendering results.

\textbf{Shading factor.} 
Due to self-occlusion, relative position to the light, and camera exposure time, the same point on a surface may have varying levels of illumination when a performer is continuously acting. Without the pose-related shading factor, the color decoder cannot adequately represent the differences between frames, resulting in a slight performance drop for both novel view and novel pose synthesis.

\textbf{Signed height indicator.} 
To investigate the influence of the signed height indicator, we set them to constant zeros during training. Without the aid of indicators, the network struggled to differentiate between diffused spatial features that could be the same at distinct positions. This resulted in artifacts along lines perpendicular to the surface and a significant drop in SSIM.

\textbf{NN interpolation.} 
As a viable alternative, we employ a self-attention block to integrate the nearest latent codes instead of relying on NN interpolation. These learnable weights possess more flexibility than inverse distance weights and can achieve comparable performance in novel view synthesis.
Nevertheless, consistency cannot be guaranteed when the positions of anchor nodes shift. Consequently, the integrated latent code of a given point may not accurately reflect the motion of the anchor nodes in a predictive manner, which is crucial for geometry editing.

\subsection{Limitation and Discussion}

Our method struggles with complex body motions due to the limited capacity of the NN skinning field. Recent works have employed root-finding-based deformation fields to align the target space with the canonical space and enhance the generalization ability of the reconstructed avatar significantly \cite{chenSNARFDifferentiableForward2021, chenFastSNARFFastDeformer2022}. Therefore, it is worth trying to replace the NN skinning with a root-finding module to further enhance the capacity of UVA. Another limitation is the fixed number of anchor nodes, which may cause artifacts when the mesh vertexes are pushed away or pulled together during editing. 
As shown in the second row of Figure \ref{fig:geo_editing}, the anchor nodes become sparse in the belly region, resulting in black stripes. This issue can be mitigated by dynamically adjusting the nodes  \cite{ruckert_AdopApproximateDifferentiable_2022}, \eg, adding more in sparse regions and removing redundant ones. We leave it as the future work.

\input{tables/ablation}

\section{Conclusion}

We present a novel approach for volumetric avatar reconstruction that enables local geometry and texture editing, high-quality novel view synthesis, and novel pose rendering. Our method incorporates structured latent codes that are attached to a deformable mesh to track the deformation of the anchored mesh to guide the geometry editing. It also enables local texture editing by spatial-aware latent code fine-tuning. Utilizing volumetric rendering, it can produce high-quality images in novel views, and the reconstructed volumetric avatar can be reposed via the skinning motion field. We believe that our method sets a baseline for all-in-one avatar models that can be rendered, reposed, and edited.

{\small
\bibliographystyle{ieee_fullname}
\bibliography{UVA,UVA_p2}
}

\end{document}

%% file: fig_tex/first_fig.tex
\begin{figure*}[t]
    \centering
    \setlength\tabcolsep{1.0pt} 
    \begin{tabular}{ c  c c c c c c}
        \includegraphics[width=0.138\textwidth]{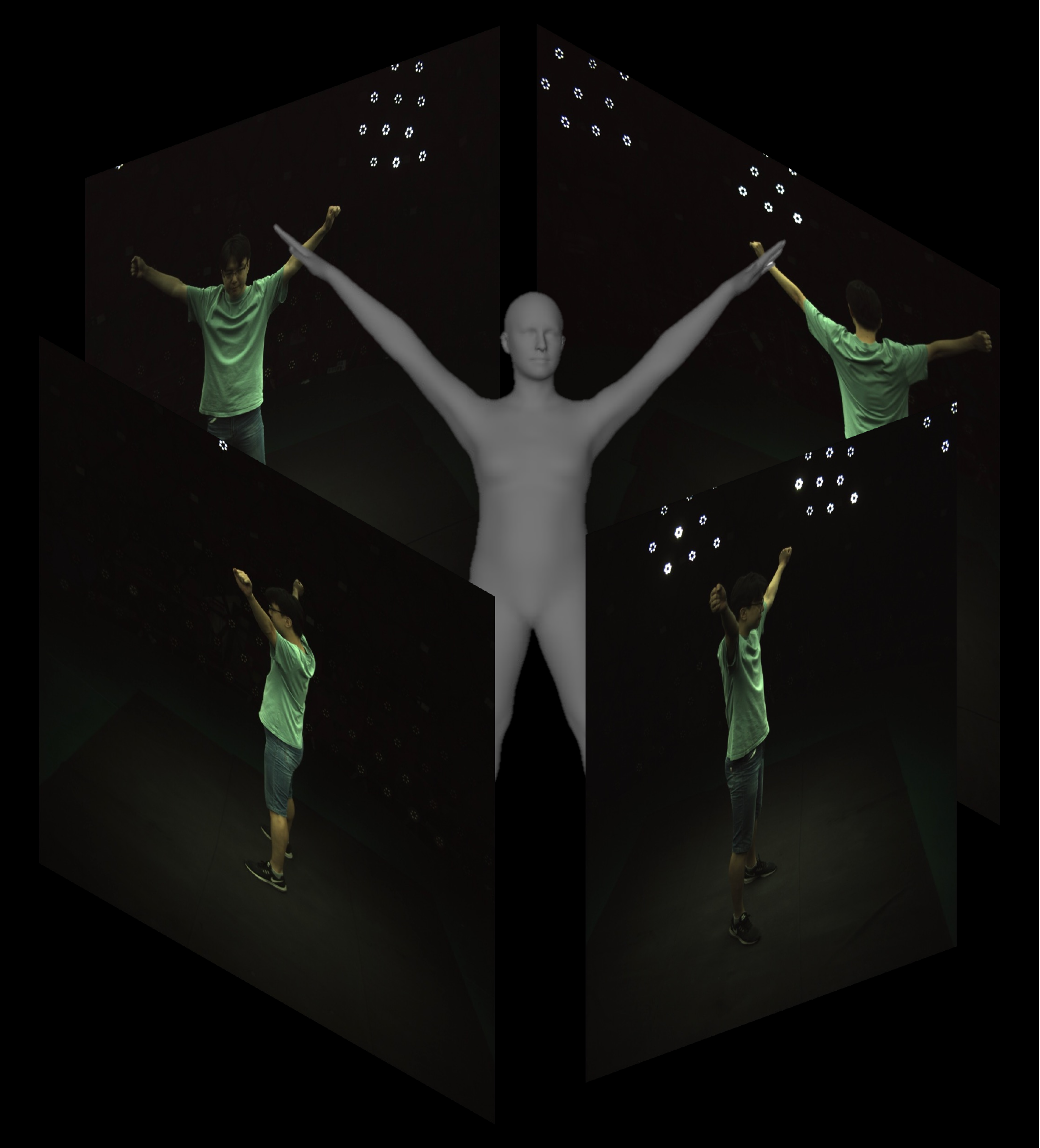} & 

        \animategraphics[controls,buttonsize=.7em,autoplay,loop,palindrome,poster=6,width=0.14\textwidth]{10}{./figures/reconstruction/pose_crop_}{0}{11} &
        \animategraphics[controls,buttonsize=.7em,autoplay,loop,palindrome,poster=5,width=0.141\textwidth]{10}{./figures/novel_view/novel_view_crop_}{0}{9} &
        \animategraphics[controls,buttonsize=.7em,autoplay,loop,palindrome,poster=last,width=0.141\textwidth]{5}{./figures/novel_pose/novel_pose_crop_}{2}{12} &
        
        \animategraphics[controls,buttonsize=.7em,autoplay,loop,palindrome,poster=last,width=0.14\textwidth]{15}{./figures/geo_gif/texture_swap_crop_}{0}{14} &
        \animategraphics[controls,buttonsize=.7em,autoplay,loop,palindrome,poster=last,width=0.14\textwidth]{2}{./figures/texture_swap/texture_}{0}{1} &
        \animategraphics[controls,buttonsize=.7em,autoplay,loop,palindrome,poster=last,width=0.14\textwidth]{6}{./figures/texture_painting/painting_test_}{0}{2} \\
        Input & Reconstruction & Novel view  & Novel pose & Geometry editing & Texture swapping & Texture painting
    \end{tabular}

    \caption{
    We reconstruct a unified volumetric avatar from a sparse multi-view video. With camera and pose parameters, we can render new views and poses of the performer. Our approach uses disentangled neural fields based on structured latent codes to capture the 3D geometry and appearance, allowing independent and local geometry and texture editing. To see the animation, please check out the documents with a compatible software like \textit{Adobe Acrobat} or \textit{KDE Okular}. We also provide the animation as a separate file in the supplementary material.
    }
    \label{fig:first_img}
    
\end{figure*}

%% file: tables/diff_check.tex
\begin{table}[!t]
\begin{center}

\begin{tabular}{l c c c c}
\toprule
Methods & NV & NP & Local Tex.  & Local Geo. \\
\hline
NeRF~\cite{mildenhallNeRFRepresentingScenes2020a} & \cmark & \xmark & \xmark & \xmark \\
Neumesh~\cite{yangNeuMeshLearningDisentangled2022} & \cmark & \xmark & \cmark & \cmark \\
Ani-NeRF~\cite{pengAnimatableNeuralRadiance2021} & \cmark & \cmark & \xmark & \xmark \\
UV-Volumes~\cite{chenUVVolumesRealtime2022} & \cmark & \cmark & \cmark & \xmark  \\
\hline
Ours (UVA) & \cmark & \cmark & \cmark & \cmark \\
\bottomrule

\end{tabular}
\end{center}
\caption{
\textit{Comparison of the rendering and editing abilities of different methods.} Our UVA enables local and independent editing of both \textbf{Geo}metry and \textbf{Tex}ture while retaining the ability to render \textbf{N}ovel \textbf{V}iews and \textbf{N}ovel \textbf{P}oses.
}
\label{tab:check_check}
\vspace*{-2em}
\end{table}

%% file: fig_tex/overview.tex
\begin{figure*}[htbp]
    \centering
    \includegraphics[width=0.9\linewidth]{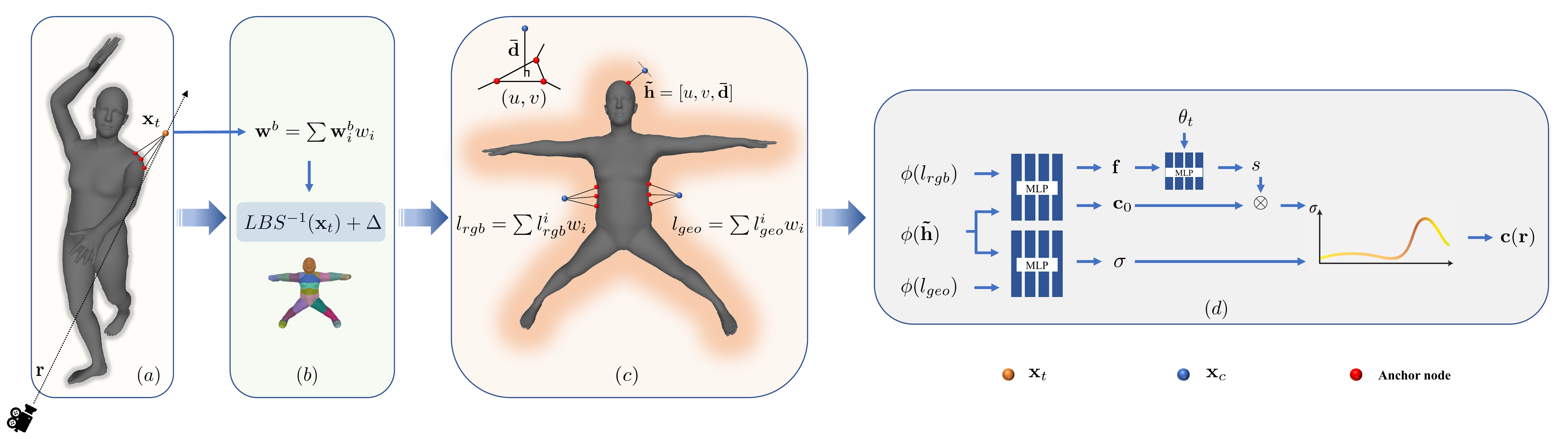}
    \caption{
    \textit{Overview of the proposed method.} Given a sampled point $\mathbf{x}_t$ in the target space (a), we first determine its LBS (Linear Blending Skinning) weights $\mathbf{w}^b$ by interpolating the weights of its Top-K nearest neighbors (NNs). Combining a neural deformation field with the NN motion field (b), the target point $\mathbf{x}_t$ is transformed into the canonical space (c). Both the structured geometry latent codes $l_{geo}$ and appearance latent codes $l_{rgb}$ are anchored to nodes on the deformable canonical mesh $\mathcal{M}$. A latent code for an arbitrary point $\mathbf{x}_c$ is then calculated through NN interpolation. Given the signed height indicator $\mathbf{\Vec{h}}$,  we use an MLP to predict the volume density $\sigma$ and color $\mathbf{c}_0$. After being modulated by the shading factor $s$, the colors are integrated along the target ray to produce the final color $\mathbf{c}(\mathbf{r})$.}
    \label{fig:overview}
\end{figure*}

%% file: fig_tex/geo_edit_313.tex
\begin{figure*}[htbp]
    \centering
    \includegraphics[width=0.7\linewidth]{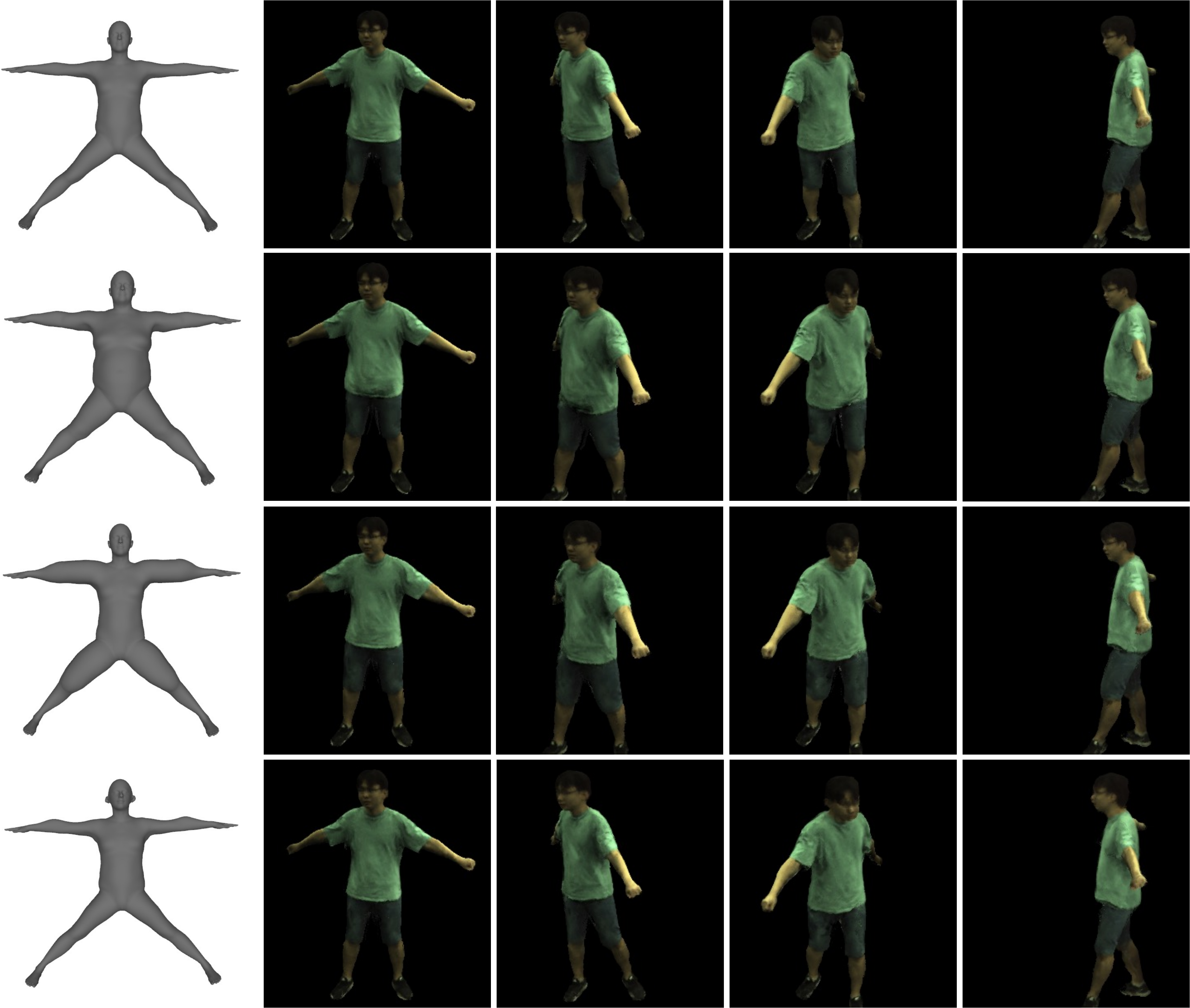}
    \caption{
    \textit{Visual results} of geometry editing. The first column shows the edited geometry in canonical pose, while the remaining columns show the rendering results from different poses. The images in the first row are rendered in reconstructed geometry. In the second row, the geometry is manipulated globally by adjusting the SMPL shape parameters to increase its size. Local geometry editing is shown in the last two rows, where the size of limbs is partly increased in the third row, and the face, ears, and arms are stretched out slightly in the last row.
    } 
    \label{fig:geo_editing}
\end{figure*}

%% file: tables/nv_np_zju_avr.tex
\begin{table}[tb]
\centering
\resizebox{1.0\linewidth}{!}{
\begin{tabular}{c cccccc}
\toprule
\multicolumn{1}{c}{\multirow{2}{*}{Methods}} & \multicolumn{3}{c}{Novel View} & \multicolumn{3}{c}{Nove Pose} \\ \cmidrule(lr){2-4} \cmidrule(lr){5-7}  
\multicolumn{1}{c}{} & \multicolumn{1}{l}{PSNR $\uparrow$} & \multicolumn{1}{l}{SSIM $\uparrow$} & \multicolumn{1}{l}{LPIPS $\downarrow$} & \multicolumn{1}{l}{PSNR $\uparrow$} & \multicolumn{1}{l}{SSIM $\uparrow$} & \multicolumn{1}{l}{LPIPS $\downarrow$} \\ \hline

NB~\cite{pengNeuralBodyImplicit2021} & 28.3 & 0.946 & 0.095 & 23.8 & 0.897 & 0.142 \\
Ani-NeRF~\cite{pengAnimatableNeuralRadiance2021} & 26.1 & 0.921 & 0.139 & 23.3 & 0.891 & 0.159 \\
A-NeRF~\cite{suAnerfArticulatedNeural2021} & 27.4 & 0.937 & 0.101 & 22.4 & 0.862 & 0.199 \\
ARAH~\cite{wangArahAnimatableVolume2022} & 28.5 & 0.948 & 0.081 & 24.6 & 0.911 & 0.107 \\
TAVA-NN~\cite{liTavaTemplatefreeAnimatable2022} & 26.3 & 0.924 & 0.112 & 23.7 & 0.892 & 0.139 \\
Ours & 26.7 & 0.924 & 0.110 & 23.9 & 0.892 & 0.137 \\

\bottomrule
\end{tabular}
}
\caption{\textbf{Quantitative results} for novel view synthesis and novel pose rendering. The metrics have been averaged over nine performers from the ZJU-MoCap dataset.
}
\label{tab:aver_metrics}

\vspace*{-2em}

\end{table}

%% file: fig_tex/nv_zju.tex
\begin{figure}[htbp]
    \centering
    \includegraphics[width=0.8\linewidth]{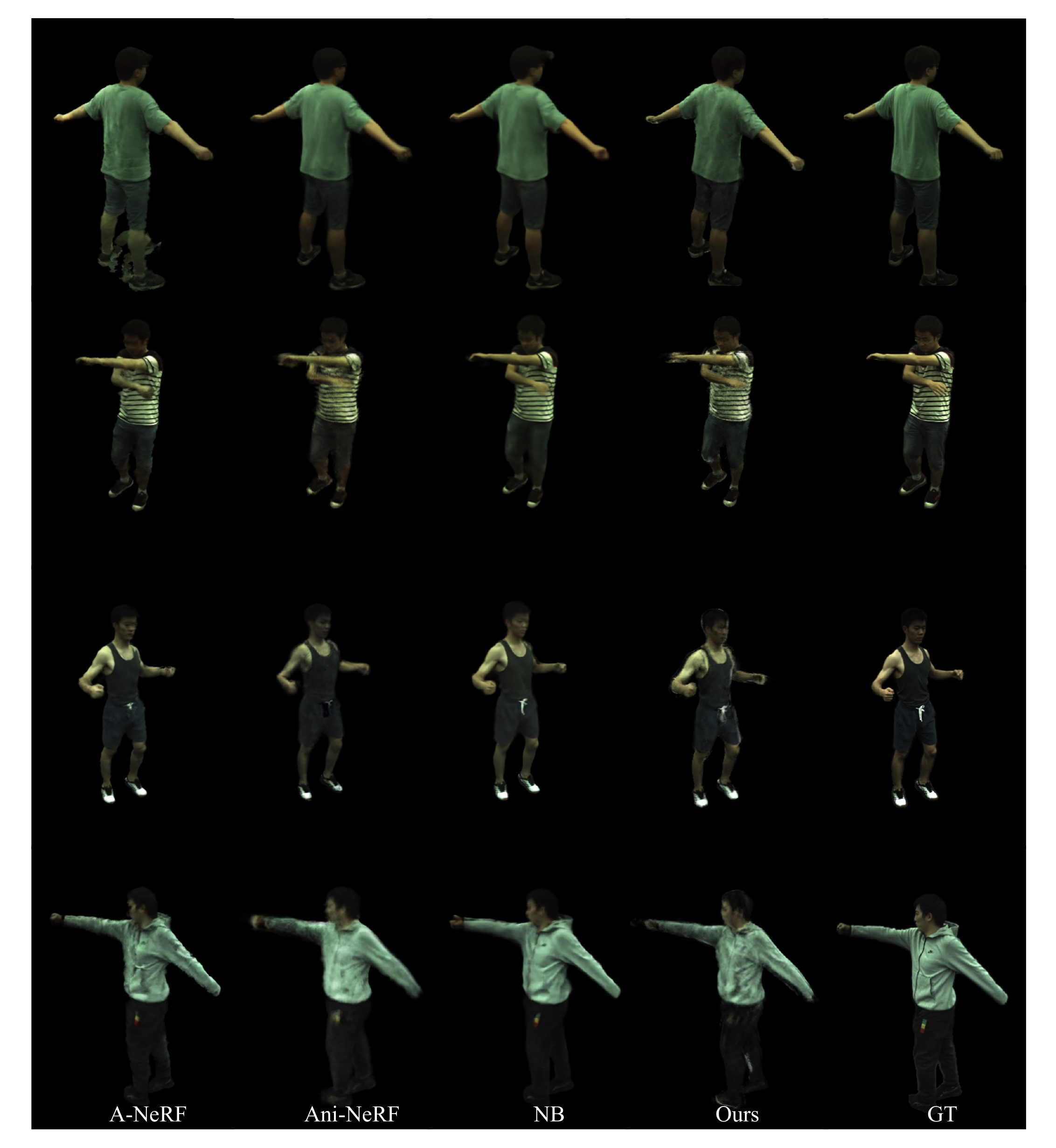}
    \caption{\textit{Qualitative results} for novel view synthesis. }
    \label{fig:nv_zju}
\end{figure}

%% file: fig_tex/np_zju.tex
\begin{figure}[htbp]
    \centering
    \includegraphics[width=0.8\linewidth]{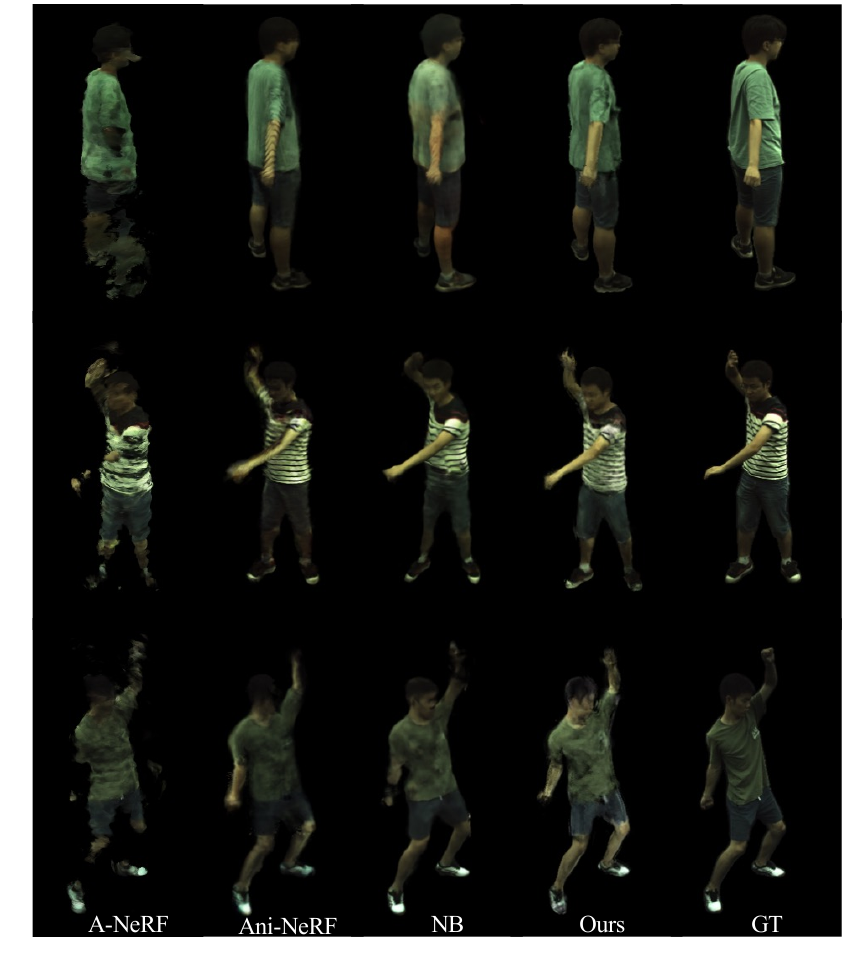}
    \caption{\textit{Qualitative results} for novel pose synthesis. }
    \label{fig:np_zju}
\end{figure}

%% file: fig_tex/edit.tex
\begin{figure}[htbp]
    \centering
    \includegraphics[width=0.95\linewidth]{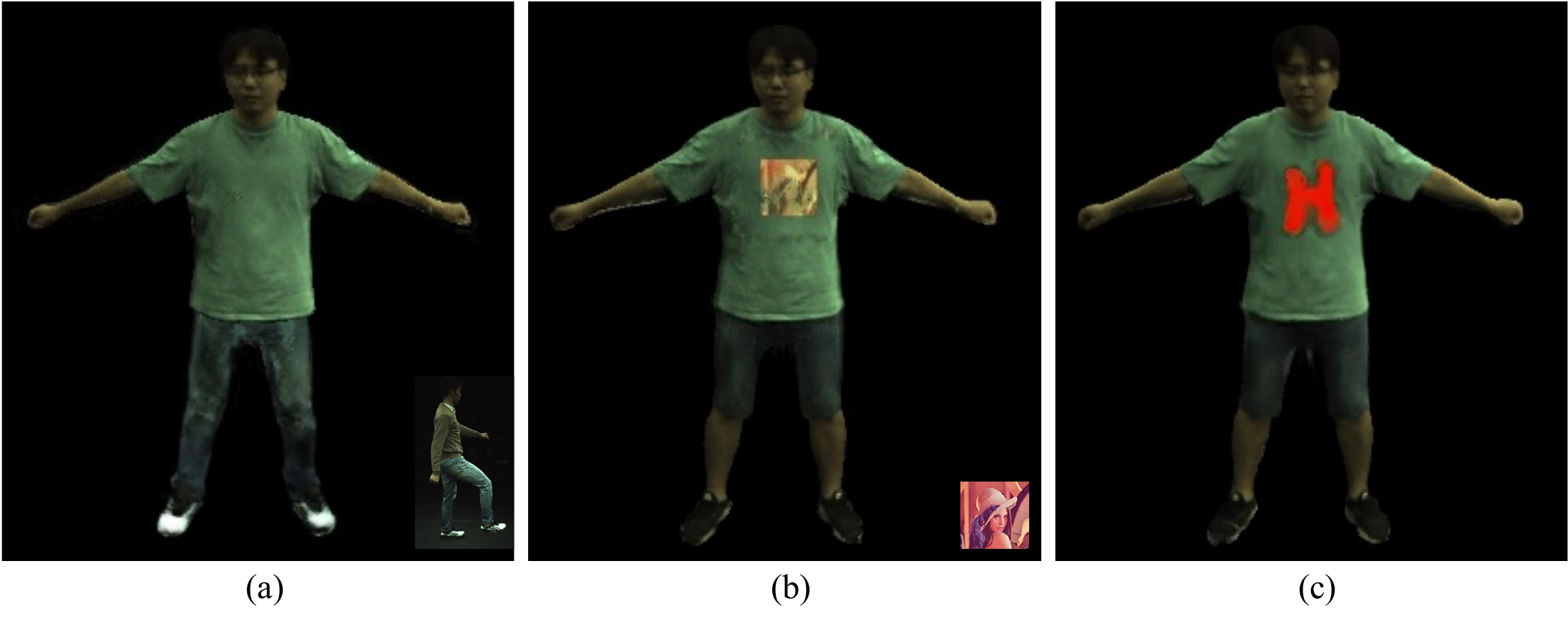}
    \caption{\textit{Visual results} of texture editing. From left to right, the results are texture swapping (a), texture painting using a second image (b), and a pattern drawing by hand (c).}
    \label{fig:editing}
\end{figure}

%% file: tables/ablation.tex
\begin{table}[t!]
\centering
\resizebox{1.0\linewidth}{!}{
\begin{tabular}{l cc cc}
\toprule
\multicolumn{1}{c}{\multirow{2}{*}{}} & \multicolumn{2}{c}{Novel View} & \multicolumn{2}{c}{Novel Pose} \\ 
\cmidrule(lr){2-3} \cmidrule(lr){4-5}
\multicolumn{1}{c}{} & \multicolumn{1}{l}{PSNR $\uparrow$} & \multicolumn{1}{l}{SSIM $\uparrow$} & \multicolumn{1}{l}{PSNR $\uparrow$} & \multicolumn{1}{l}{SSIM $\uparrow$} \\ \hline

w/o $\Delta$ & 30.2 & 0.958 & 23.9 & 0.903  \\
w/o shading & 29.0 & 0.953 & 24.1 & 0.902  \\
w/o indicator & 29.3 & 0.945 & 24.0 & 0.894  \\
Att. fusion & 30.4 & 0.957  & 23.9& 0.902  \\
\hline
Ours(UVA) & \textbf{30.5} & \textbf{0.958}  & \textbf{24.5} & \textbf{0.905}  \\
\bottomrule
\end{tabular}
}

\caption{
\textit{Abalation study} of design choices in our method.
}
\label{tab:ablation}
\vspace*{-2em}
\end{table}